\documentclass[10pt,journal,compsoc]{IEEEtran}
\usepackage{amsmath}
\usepackage{booktabs}
\usepackage{subfigure}

\ifCLASSOPTIONcompsoc
  \usepackage{algorithm}
  \usepackage[noend]{algpseudocode}
\else
  \usepackage{cite}
\fi

\algdef{SE}[DOWHILE]{Do}{doWhile}{\algorithmicdo}[1]{\algorithmicwhile\ #1}%
\ifCLASSINFOpdf
\else
\fi
\usepackage{graphicx}
\graphicspath{ {./resources/} }
\begin{document}
%
\title{AffectON: Incorporating Affect Into Dialog Generation}
%
%
%
%

\author{Zana~Bu\c cinca,
        Y\"{u}cel~Yemez,
        Engin~Erzin,
        Metin~Sezgin
        
\thanks{Z. Bu\c cinca is with the School of Engineering and Applied Sciences, Harvard University, Cambridge MA, USA.\protect\\
Y. Yemez and M. Sezgin are with the KUIS AI Lab and Computer Engineering Department, Ko\c{c} University, Istanbul, Turkey.\protect\\
E. Erzin is with the KUIS AI Lab, Computer Engineering Department and Electrical \& Electronics Engineering Department, Ko\c{c} University, Istanbul, Turkey. \protect\\
E-mail: zbucinca@seas.harvard.edu, \{yyemez, eerzin, mtsezgin\}@ku.edu.tr}
}

%
%

\markboth{IEEE TRANSACTIONS ON AFFECTIVE COMPUTING, VOL. 11, NO. 3, JULY-SEPTEMBER 2020}%
{Shell \MakeLowercase{\textit{et al.}}: Bare Demo of IEEEtran.cls for Computer Society Journals}
%



\IEEEtitleabstractindextext{%
\begin{abstract}
Due to its expressivity, natural language is paramount for explicit and implicit affective state communication among humans. The same linguistic inquiry (e.g. \textit{How are you?}) might induce responses with different affects depending on the affective state of the conversational partner(s) and the context of the conversation. Yet, most dialog systems do not consider affect as constitutive aspect of response generation. In this paper, we introduce \textit{AffectON}, an approach for generating affective responses during inference. For generating language in a targeted affect, our approach leverages a probabilistic language model and an affective space.  \textit{AffectON} is language model agnostic, since it can work with probabilities generated by any language model (e.g., sequence-to-sequence models, neural language models, n-grams). Hence, it can be employed for both affective dialog and affective language generation. We experimented with affective dialog generation and evaluated the generated text objectively and subjectively. For the subjective part of the evaluation, we designed a custom user interface for rating and provided recommendations for the design of such interfaces. The results, both subjective and objective demonstrate that our approach is successful in pulling the generated language toward the targeted affect, with little sacrifice in syntactic coherence.
\end{abstract}

}

\maketitle

\IEEEdisplaynontitleabstractindextext

%
\IEEEpeerreviewmaketitle

\IEEEraisesectionheading{\section{Introduction}\label{sec:introduction}}

%
%
%
%
\IEEEPARstart {C}{onsidered} as the foundation of civilization, language is the most sophisticated and intuitive means of communication amongst humans. With their utterances, along with exchanging information and expressing ideas, people explicitly or implicitly indicate their affective states.  Yet, we are surrounded by computers that for a long time could interpret commands and convey outputs solely in an ``unnatural'' language - code.  This gulf between verbal human and computer communication has been vastly bridged by advances in natural language processing, resulting in emerging interfaces such as conversational agents and dialog systems. Nevertheless, as Picard \cite{picard2002computers} states, for genuinely natural and intelligent human-computer interaction, computers should have and express emotions. 

Previous research shows that emotionally responsive (compared to non-responsive) conversational agents are rated more positively and perceived more helpful and trustworthy by users \cite{kuhnlenz2013increasing, lutfi2013feel}, which in turn contributes to more efficient human-computer teams. Moreover, evidence suggests that affective conversational agents can be helpful in an array of applications. They assist students in learning efficiently \cite{kort2001affective}, alleviate mental health conditions (e.g. depression \cite{taylor2017personalized}, dealing with bullying \cite{GordonLBKFLHT19}), show empathy when delivering emotional information \cite{mahamood2011generating}, provide medical care humanely \cite{malhotra2015exploratory} and improve the overall emotional well-being \cite{ghandeharioun2019emma}. In this context, while the existing systems are increasingly aspiring for more naturalistic and humane interfaces, they still lag behind on the incumbent ability of affect generation that prior research has shown to have positive impact on users.

Affect is defined as a set of observable manifestations of a subjectively experienced emotion \cite{Dict}. Being an intrinsic aspect of humans, it is exhibited through visual, vocal and verbal cues. To illustrate, people reveal their joy by their happy facial and vocal expressions, but also through their rich-in-positive-words utterances. A plethora of studies have delved into synthesizing affect in computers (i.e. agents or robots) visually by facial expressions and body language \cite{andre2000integrating, huang2017dyadgan, hudlicka2016virtual}, or vocally by speech features \cite{pierre2003production, henton1999method}, \cite{adigwe2018emotional}.  On the other hand, not as much effort has been spent on the equally vital task of affective language generation.

Following Munezero et al.'s definition \cite{munezero2014they}, in this study, affective language generation is deemed different from sentimental or emotional language generation. Although in the literature, there persists a lack of lucid distinction among affect, emotion, feeling and sentiment terms, as they are frequently used interchangeably, affect is unanimously considered as the term that subsumes emotion, feeling and sentiment \cite{fleckenstein1991defining}. The three principal dimensions of affect consist \textit{valence} (pleasant - unpleasant), \textit{arousal} (active - passive), and \textit{dominance} (dominant - submissive) \cite{russell2003core}. Evidently, comprised of three dimensions, affect is a continuous multifaceted phenomenon, in contrast to sentiment which is gauged solely by polarity (\textit{positive -- negative}), or emotion which has discrete categories (e.g. \textit{happy, sad, angry, afraid, disgust}). 

Generating affective language adds further complexity to the already challenging task of natural language generation. Besides meeting the natural language requirements in terms of semantics, syntax and morphology \cite{de1996architectures}, the text should also convey the targeted affect. If the generated language, in addition to these requirements, is also conditioned on another sentence, the task transforms into affective dialog generation. As a result, affective language generation can be regarded as a subproblem of affective dialog generation.  

In this paper, we present \textit{AffectON}, an approach for affective dialog generation, which can also be applied to the affective language generation setting. \textit{AffectON} is a language model agnostic algorithm which generates natural language in a targeted affect during inference by leveraging a probabilistic language model and an affective space. Since affective dialog generation subsumes affective language generation, the focus of the paper is affective dialog generation. Though \textit{AffectON} is compatible with any probabilistic language model, in this study we experiment with a sequence-to-sequence neural language model for affective dialog generation. We evaluate the generated responses objectively, in terms of \textit{valence} and \textit{perplexity}. For subjective evaluation, we design a custom user interface for rating the generated text in terms of \textit{valence, arousal, dominance, syntax}, and \textit{appropriateness}. The results indicate that our approach is successful in generating responses close to the targeted affect, while preserving the syntactic and semantic requirements.

Our contributions can be summarized as follows:
\begin{itemize}
	\item We present an approach for affective dialog and language generation by leveraging a language model and an affective space. To the best of our knowledge, we introduce the first language model agnostic affective language generation approach, which can be employed by any underlying language model.
    \item We design a protocol and a user interface for subjective evaluation of affective text generation systems and provide design recommendations for such interfaces.
\end{itemize}

\section{Related Work}

The main lines of research that tackle affective language generation fall under the categories of textual style transfer and affective dialog systems. The relevant related work in these areas is discussed below.

\subsection{Textual Style Transfer}

Style transfer is regarded as the task of generating semantic content in a targeted style. Recent advances in deep learning have led to a surge of research on style transfer via deep neural nets. While the task has enticed chiefly the computer vision community, with multiple works achieving remarkable results in style transfer on images \cite{ Gatys_2016_CVPR, karras2018stylebased, isola2017image}, far fewer studies exist in language style transfer. Akin to image style transfer, language style transfer rephrases a style-free content with the desired stylistic attributes. For instance, Jhamtani et. al \cite{jhamtani2017shakespearizing} transfer phrases from standard English to Shakespearean English. They utilize parallel corpora and an encoder-decoder network for training.

Though Tikhonov and Yamshchikov argue that sentiment is not a stylistic attribute of language \cite{ tikhonov2018wrong},  many studies consider sentiment to be so, hence manipulate with it accordingly. For instance, Hu et. al \cite{ hu2017toward} aim to generate controllable sentences by learning disentangled latent representations. They leverage variational auto-encoders, while constraining generation based on specified attributes, such as sentiment or tense. Shen et al. \cite{shen2017style} also generate sentences in a targeted sentiment by separating semantic and stylistic content of the text. They train an encoder that extracts style from the source text, yielding a style-free latent representation, which is then fed to a decoder together with a targeted style.  Wang et. al \cite{Wang2018SentiGANGS} mix multiple generators for sentimental text generation. Each generator is responsible for learning to generate sentences with a single sentiment, but they all go through one discriminator.

By modifying sequence-to-sequence architectures, Fu et al. \cite{ fu2018style} propose two models for style transfer. In the first one, they encode the source sentence into a latent representation, by removing the stylistic information. Afterwards, they feed the content to two separate decoders which endow the content with predefined styles (sentiments). The other model utilizes a single decoder, which on top of content is conditioned on style also.
Recently, pre-trained large transformer based models have been used to condition the sentiment of the generated text \cite{Dathathri2020Plug}.
Our work is similar to the studies discussed here as it also transfers the affect of the sentences. Nonetheless, we are interested in the whole affect of the sentence, not only the sentiment, which is confined to \textit{positive} or \textit{negative}. In contrast to these studies, we do not regard affect as a stylistic attribute of the text. In addition, our approach is post-hoc, as we do not update model weights when transferring the affect of the sentence.

\subsection{Affective Dialog Systems }

Research on affective dialog system can be broadly classified into rule-based and data-driven approaches.
In rule-based approaches, hand-crafted rules are embedded into the dialog systems to make them more affective. These approaches are specific to the domain and cannot be employed in open domain settings. For instance, Mahamood et al. \cite{mahamood2011generating} generate affective text for systems providing information to parents of neonatal infants. 

Advances in deep learning gave impetus to data-driven dialog modeling. Vinyals et al. \cite{vinyals2015neural} showed that relative success in data-driven conversation modeling could be achieved with end-to-end training of sequence-to-sequence frameworks. Nonetheless, these approaches entail shortcomings such as producing short, dull and emotionless answers. To obtain affectively richer responses, a couple of studies have incorporated emotion into dialog generation.
Zhou et. al \cite{ zhou2018emotional} introduce a modified sequence-to-sequence architecture that produces answers, conditioned on one of eight emotion categories (angry, disgust, fear, like, happy, sad, surprise, other). Their architecture is comprised of an external and internal memory. The external memory is responsible for deciding whether to choose an emotional or non-emotional word during decoding. Whereas, the internal memory unit controls the amount of the emotion expressed during the decoding. Another study \cite{huang2018automatic} that conditions responses on emotion categories, experiments with the ways of concatenating emotion tokens with the latent representation of the source utterance. Although, the results did not show a huge difference among the ways of concatenation, they observed that for some emotion categories (i.e. \textit{fear}) concatenating emotion token prior to the context vector yielded marginally better results. The aforementioned studies are relatively successful in generating emotionally charged responses. However, the treatment of emotions as discrete categories makes these approaches impractical for conversation modelling, since the conversation would have abrupt, and unnatural changes in terms of emotion. Meanwhile, following the work of \cite{bradley1999affective}, we regard affect as a continuous, multidimension phenomenon while generating responses. This in turn enables smooth transitions among affective states.

Asghar et al. \cite{asghar2018affective} utilize a continuous affective space and sequence-to-sequence framework for affective response generation. To boost the performance, they utilize affective embeddings during encoding, augment the loss function with an affective objective and conduct affectively diverse beam search. Their approach does not require a target affect, instead the affective direction is guided by the loss function. They experimented with three distinct strategies for the loss function. The first strategy generates responses in the same affective direction with the source sentence. The second one generates responses in the opposite affective direction with the source sentence, and the final strategy tries to generate non-neutral responses irrespective of the direction. While these strategies might be useful in certain scenarios, we find them unnatural since humans do not have a pre-decided affective direction strategy when conversing. Instead, our responses and their affective direction develop naturally depended on multiple factors such as: the context of the conversation, our current mood and the relationship with the conversational partner. Thus, we argue that for naturalistic conversations the responses should be conditioned on affective information. Hence, in contrast to this study our approach generates responses conditioned on target affects. Moreover, we do not include affective information while training the seqeuence to sequence model, but utilize the learned weights with external affective information for response generation.

More recently and also closely related to our work, Colombo et al. \cite{colombo2019affect} propose multiple approaches to generating emotional responses. They experiment with representation of emotional content as both distribution of probabilities on a vector of discrete emotions and a continuous affective state. Among their proposed approaches, Word Level Explicit model seeks to learn generation of affective sentences by adaptively sampling affective words. The main difference from our work is that we conduct the affective mapping during inference and not training. Because we change the affect during inference, our approach is robust and can be easily incorporated into any existing language model. Further, in terms of implementation their proposed model is a GRU-based model, whereas we use a transformer based model.

\begin{figure*}[t]
\subfigure[Valence - Arousal]{
\includegraphics[width=.33\textwidth]{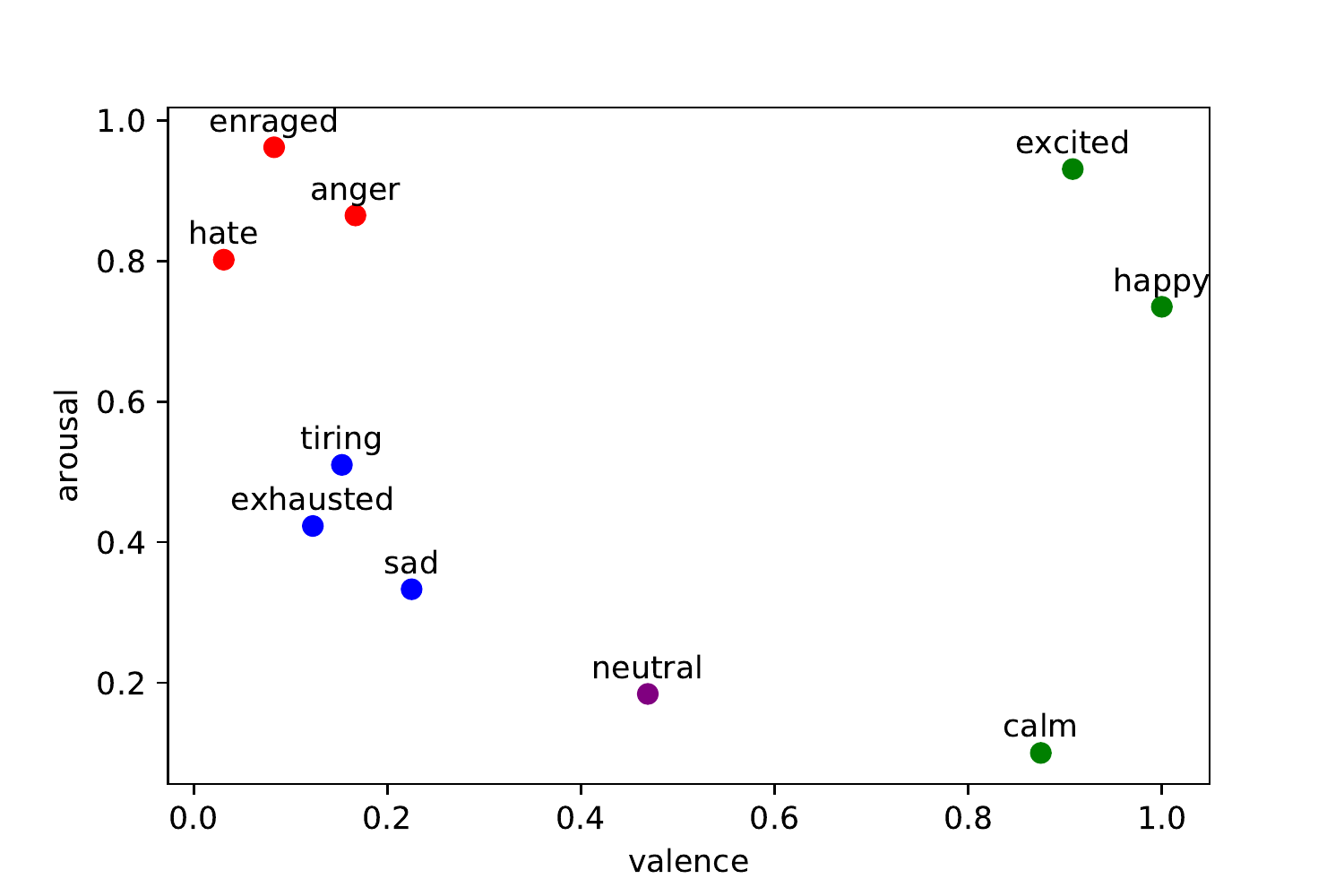}
} \subfigure[Valence - Dominance]{
\includegraphics[width=.33\textwidth]{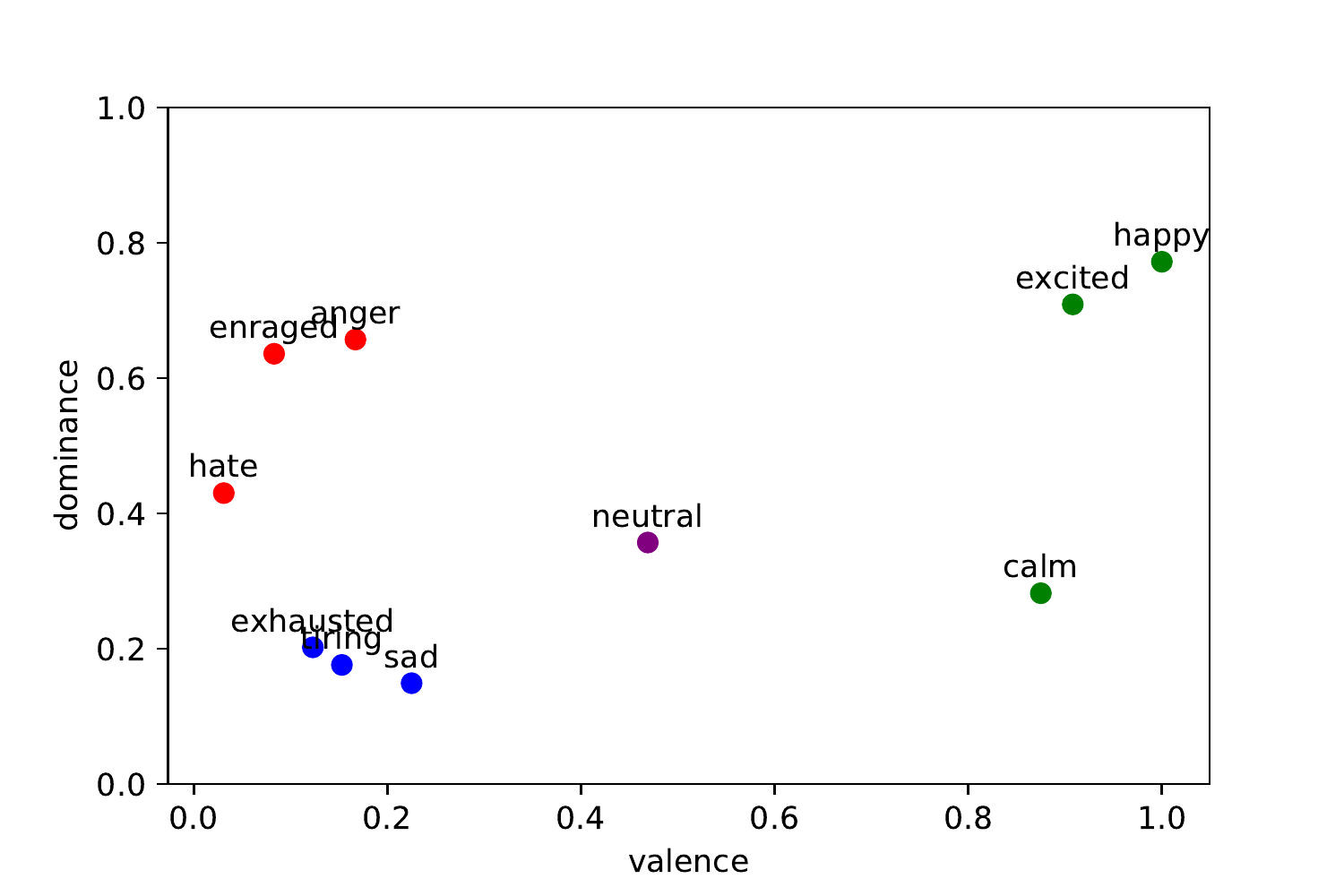}
} \subfigure[Arousal - Dominance]{
\includegraphics[width=.33\textwidth]{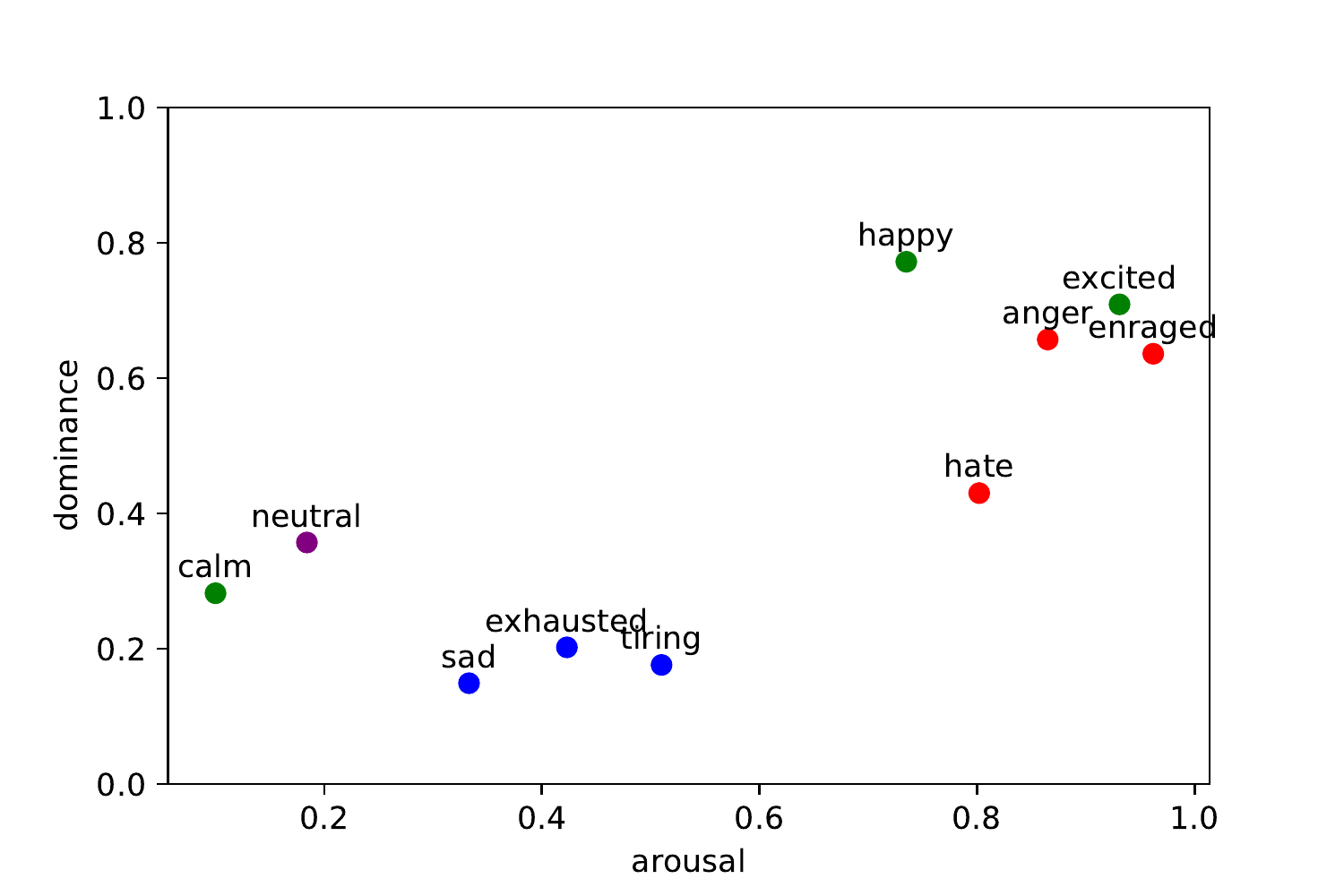}
}

\caption{Sample words in affective space \cite{mohammad2018obtaining}.} 
\label{fig:VAD}
\end{figure*}
  
\section{Background}

Prior to elaborating on the proposed approach, we will briefly describe the three principal units our method employs. 

\subsection{Language Models}

Language modeling is a core natural language processing task, applications of which span across multiple areas. A language model captures the distribution of a sequence of words in natural language. Given a sequence of words, it assigns a probability to the sequence. Essentially, we expect a language model to assign higher probabilities to sequences of words it has encountered in its training set.
By exploiting the chain rule of probability, the problem of calculating the joint probability of a sequence $N$ words $w_1, w_2,..., w_N$ is formulated as:
\begin{equation}\label{eq:prob_w}
    P(w_1,w_2,...,w_N)=\prod_{t=1}^{N} P(w_t|w_1,w_2,...,w_{t-1}).
\end{equation}

Common neural network architectures used for language modeling include the eminent Long Short-Term Memory (LSTM) architectures. LSTMs are a special type of Recurrent Neural Networks (RNNs) unfettered by the vanishing gradient problem typical RNNs suffer from \cite{hochreiter1998vanishing}. Just like RNNs, LSTMs process one token at a time and can handle variable size sequences and preserve information from predecessor words, offering a robust structure for language modeling. 
Recently, transformer models have changed the landscape in natural language processing as they have outperformed prior models in language modeling and a vast number of other end tasks \cite{vaswani2017attention, radford2018improving}. In contrast to LSTMs, these architectures generally do not employ any recurrent networks. Instead, they use the attention-mechanism at each step to decide which parts of the input sequence are relevant.

\subsubsection{Sequence-to-sequence Models}
Sequence-to-sequence models are encoder-decoder neural networks with genesis from machine translation. Specifically, these architectures are utilized to encode a sentence into one language and decode it into another. Nonetheless, apart from translation sequence-to-sequence architectures have found applications in many multimodal and unimodal learning tasks \cite{baltruvsaitis2019multimodal}. Here, we are particularly interested in their usage in conversation modeling. The idea is to encode one (or more) sentences to an embedding and then to decode this embedding to the response for the encoded utterance. For conversation modeling, sequence-to-sequence architectures employ LSTMs or transformers for both the encoder and decoder. In this manner, they provide a medium for response generation of a given utterance. Sequence-to-sequence models build upon language models, as they are conditioned on the meaning of the encoded utterance, in addition to the predecessor words of the response currently being generated. Consequently, the problem sequence-to-sequence models tackle is finding the probability of a response $Y$ = $y_1, y_2,..., y_N$, given the utterance $X$:
\begin{equation}\label{eq:prob_y}
P(Y|X)=\prod_{t=1}^{N} P(y_t|y_1,y_2,...,y_{t-1}, X).
\end{equation}

In this study we utilize a large transformer model fine-tuned on dialog corpus as a sequence-to-sequence model. Note that our approach can be implemented with any underlying language model. We chose to experiment with Generative Pre-Training (GPT) model \cite{radford2018improving} as one of the large and high performing transformer models. The pace by which new and superior language models are being released, highlights the importance of our approach being language model-agnostic and requiring no further training.

\subsection{Affective Space - VAD}

Language models capture contextual meaning of the words, but do not distinguish words according to the affect they elicit. For example, contrasting affective words, such as \textit{good} and \textit{bad}, or \textit{terrible} and \textit{wonderful} appear in similar contexts (e.g. "the weather was terrible" and "the weather was wonderful") and are given similar probabilities under the language model. An affective space, where words similar in affect are proximate, would be valuable for analyzing and comparing words according to their underlying affect.

ANEW \cite{bradley1999affective}  is one of the earliest lexicons to consider affective meaning of words in three dimensions. In this study, humans rated 1034 words on a scale from 1 to 9 in terms of valence, arousal and dominance. This affective view accounts \textit{valence} to range from unpleasant to pleasant, \textit{arousal} to range from calm to excited and \textit{dominance} from submissive to dominant. Following the same approach, Warriner et al. \cite{warriner2013norms} provide an extended version of ANEW lexicon with 13,915 lemmas rated in valence, arousal and dominance (VAD) dimensions.  

More recently, Mohammad \cite{mohammad2018obtaining} introduced \emph{NRC Valence,
Arousal, and Dominance (VAD) Lexicon}, which contains 20,007 VAD values for lemmas rated on a continuous scale from 0 (low) to 1 (high). These ratings were obtained by using \emph{Best-Worst Scaling}, as an approach for overcoming the shortcomings of \emph{Likert} scales (e.g. biases towards the middle of the scale) that were used for obtaining the ratings in prior work.  We utilize this lexicon as an affective space.
Figure~\ref{fig:VAD} depicts sample words in VAD space. 


\begin{figure*}[t]
\subfigure[Traditional encoding-decoding for dialog generation]{
\includegraphics[width=\linewidth]{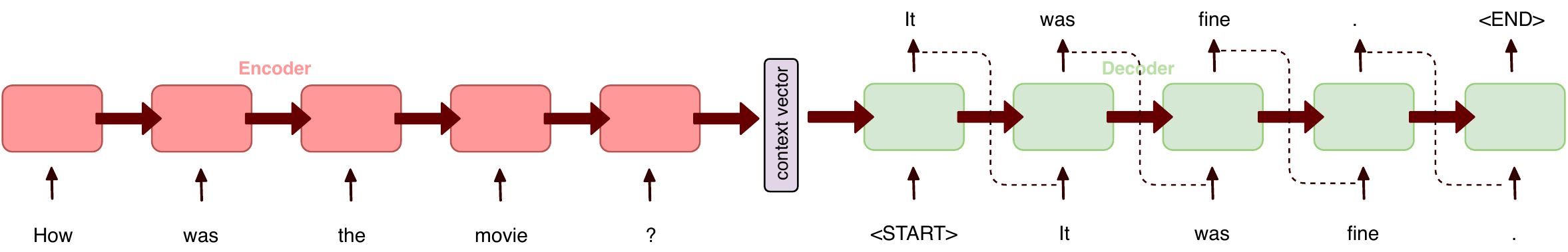}
} \subfigure[Decoding with AffectON]{
\includegraphics[width=\linewidth]{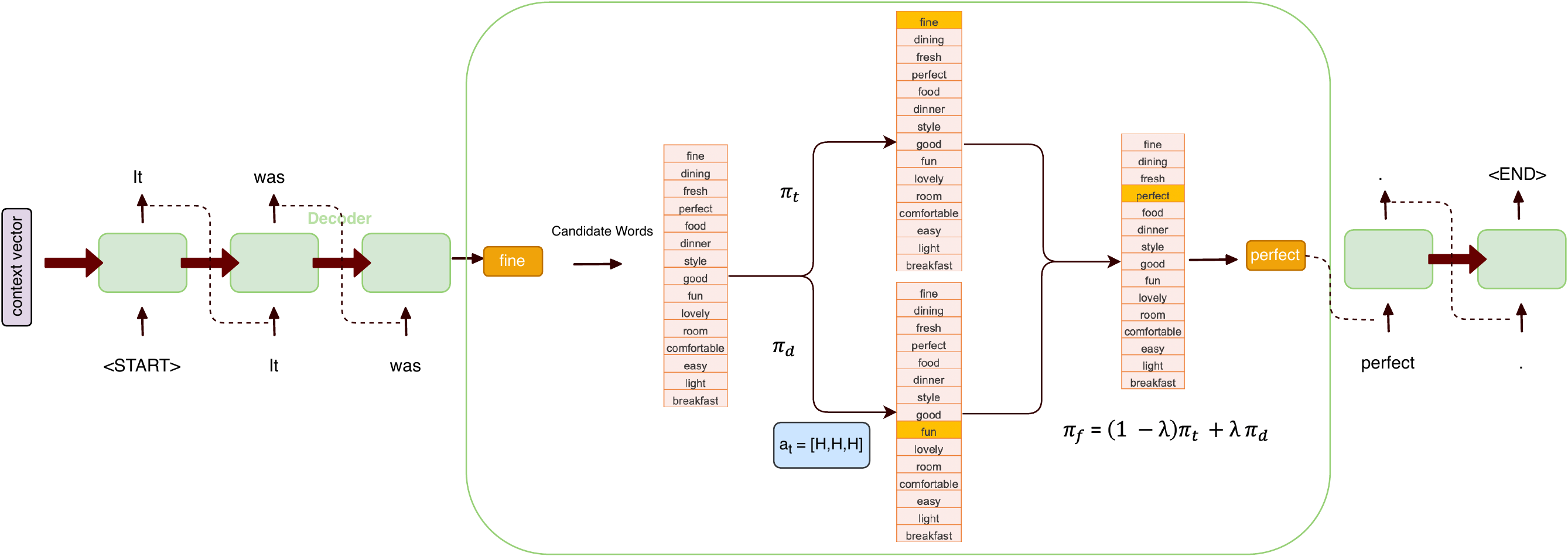}
}
\caption{ Traditional encoding-decoding (a) and the intervention of AffectON in decoding (b) are shown. Subfigure (b) depicts an example run through a simplified pipeline of AffectON with a sequence-to-sequence model. The highlighted words in the lists of words represent the words with the highest probability in that particular probability distribution.}
\label{fig:system}
\end{figure*}

\section{Proposed Method: AffectON}
\label{AffectON}

In this section, we describe the proposed approach for text generation in a target affect. Specifically, we elaborate on affective response generation with a sequence-to-sequence model. However, our approach can be employed by any probabilistic language model. 

\subsection{Problem Formulation}
Let $s_t$ be a sequence of words ${w_1, w_2, ..., w_t}$  that is semantically meaningful, syntactically correct and corresponds to the target affect $a_t$. If we choose the next word, $w_{t+1}$ such that it also belongs to the target affect $a_t$ (or neutral affect $a_n$) and preserves the semantic and syntactic validness of the sequence, we will have a longer sequence $s_{t+1}=${$w_1, w_2, ..., w_t, w_{t+1}$} that also holds the required premises (i.e., is a syntactically and semantically valid sequence in the targeted affect). The response generation scenario introduces an additional constraint, since the generated reply, in addition to being semantically meaningful, syntactically correct and in the targeted affect, should also be a plausible response for the given source utterance. Now, the problem transforms to the following: given a source utterance $X$, the aim is to generate a response $Y$ for it, in the targeted affect $a_t$.

\subsection{Candidate Words}

We use a sequence-to-sequence model comprising an encoder and a decoder, which can be either trained from scratch or fine-tuned on a dialog corpus (see section \ref{ExS} for implementation details). The sequence-to-sequence model encodes the source utterance $X$ into a latent representation, and then decodes it to obtain the response $Y$. Thus, the model has learned not only the underlying language distribution, but also appropriate response generation. 

To generate these responses in the targeted affect $a_t$, we encode the source utterance $X$ into a latent representation $h$. We intervene with the decoding part. Figure~\ref{fig:system} depicts the modification of the affect of a generated word with AffectON. At any given step $t$ of the decoding, the model's output passes through a softmax function to yield the probability distribution of words $\pi_t \in {\rm I\!R^V}$, where $V$ is the size of the vocabulary. The \textit{argmax} of $\pi_t$ is essentially the index of the word $y'_t$, which sequence-to-sequence model assigns highest probability to become the next word in the response $Y$. To continue, we can either use this predicted word $y'_t$ or we can use the word $y_t$ from the ground truth response coming from the corpus.  Here, we continue with the ground truth response. We proceed by extracting the lemma form of the ground truth word $l_t = lemma(y_t)$ (e.g. $loved$ becomes $love$). This is done since the affective lexicon is constructed solely from lemmas.  We check whether the lemma at step $t$, $l_t$ exists in the affective lexicon. If the lemma is not in the affective lexicon, we do not possess any information on its affect, hence the original form of the lemma, $y’_t$ is appended to the response $Y$ without any further modification.

In the case when the lemma exists in the affective lexicon, our aim is to find a word that is contextually similar to $y’_t$, but is closer to the target affect $a_t$ than $l_t$ . To accomplish that, we pick $k$ words which the model assigns the highest probabilities as candidate words at step $t$ from the probability distribution over all words in the vocabulary $\pi_t \in {\rm I\!R^V}$. This reduced vector $\pi_t \in {\rm I\!R^k}$ of probabilities is then put through a \emph{softmax} to normalize the probabilities.

\subsection{Incorporating the Affect of Candidate Words}
To extract affective information for the candidate words, initially we obtain their lemma form. For each candidate lemma $l$, we compute the distance of its $V_lA_lD_l$ from the target affect's $a_t = V_tA_tD_t$, as Euclidean distance: $d = \sqrt{(V_l - V_t)^2 + (A_l- A_t)^2 + (D_l - D_t)^2}$, to obtain a vector of distances $d = [d_1, d_2, ..., d_k]$. To find a probability distribution over the distances, the vector of distances goes through a \textit{softmax} function to yield $\pi_d \in {\rm I\!R^k}$. Note that since a smaller distance from the target affect is more desirable, we take the negative values of the distances prior the \textit{softmax} step.
Finally, we calculate the fused probability $\pi_f \in {\rm I\!R^k}$ for each candidate word, by taking into account the probability of the language model and the distance from the target affect as a probability:
\begin{equation}\label{eq:weighted}
    \pi_f = (1 - \lambda) \pi_t + \lambda \pi_d.
\end{equation}

The parameter $\lambda$ in the equation, is tuned based on the priority we give the affect as opposed to the language model. In other words, it can be regarded as the \textit{affect strength} we aim for. If $\lambda = 1$, then the generated utterances may be closer to the target affect, but will lose in syntactic correctness and/or semantic meaningfulness. On the other hand, if $\lambda = 0$ the generated utterances will aim for preservation of the latter requirements, without any concern about being in the target affect. In other words, it will be the case of plain sequence-to-sequence decoding. Finally, the candidate word with the highest probability $\pi_f$ is picked as the next word of the response. We feed this word to the decoder, to continue the same described process for step $t+1$. Note, that if the original word was not an affective one to begin with, it is not modified, instead it is immediately fed to the decoder.

\begin{algorithm*}[!htb]
\caption{AffectON algorithm}\label{alg:euclid}
\begin{algorithmic}[1]
\State \textbf{Input:} $X, a_{t}$ \Comment{$X$ is input utterance, $a_t$ is target affect}
\State $Y'\gets \{ \}$ \Comment{set response to empty set}
\State $t \gets 0$
\State $h\gets \psi_{e}(X)$ \Comment{encode the utterance}
\While{$y'_t\neq EOS$} \Comment{while the generated word is not End Of Sentence token}
\State $t \gets t + 1$
\State $\pi_t \gets softmax(h)$ \Comment{probability distribution over the vocabulary at step $t$}
\State $y'_t \gets argmax(\pi_t)$
\State $l_t \gets lemma(y'_{t})$ \Comment {get the lemma form of the word}

\If{$l_t$ not in affective lexicon}
  \State $Y \gets Y \cup y'_t$ \Comment{add word to response if not affective}
\Else
 
  \State $c \gets top(\pi_t, k)$ \Comment{top-k words with highest probabilities}
  \State $\pi_t \gets softmax(\pi_t[c])$ \Comment{probabilities of candidate words}
  \State $l \gets lemma(c)$ \Comment{lemmas of candidate words}
  \State $d \gets euclidean(l, a_t)$ \Comment{distances of candidate words from target affect}
  \State $\pi_d \gets softmax(-d)$ \Comment{probability distribution over the distances}
  \State $\pi_f  \gets (1 - \lambda) \pi_t + \lambda \pi_d$ \Comment{aggregated scores of candidate words}
  \State $y''_t  \gets argmax(\pi_f)$ \Comment{candidate with highest score}
  \State $Y \gets Y \cup y''_t$ \Comment{add word to response}
  \State $h \gets \psi_{d}(h, y''_t)$ \Comment{feed the current picked word to the decoder}
\EndIf
\EndWhile
\State \textbf{return} $Y$\Comment{return response}
\end{algorithmic}
\end{algorithm*}

\section{Experiments}

\subsection{Model}

As large pre-trained language models are being deployed at a rapid pace, it would be useful to generate affective text by leveraging these models' generation powers either by fine-tuning them or intervening during inference. Being language model-agnostic, our proposed approach allows the utilization of any underlying model for affective text generation, thus is useful in keeping up with the pace of these developments.
We experimented with affective dialog generation by utilizing an instance of such large pre-trained model fine tuned on dialog corpora. We used an off-the-shelf model \cite{wolf2019} which won automatic metric track of the dialog competition ConvAI2 \footnote{http://convai.io/} at \textit{NeurIPS2018}. The generative model is a multi-layer Transformer encoder based on Generative Pretrained Transformer \cite{radford2018improving}. The decoder is comprised of a 12-layer decoder only transformer with masked self-attention heads (768 dimensional states and 12 attention heads). This model is fine-tuned on a dialog dataset -- \textsc{Persona-Chat} \cite{zhang2018personalizing}. \textsc{Persona-Chat} is a crowd-sourced dataset, where each speaker was asked to converse with another person conditioned on a few sentences that defined their personality. We refer the readers to the original paper \cite{wolf2019} for more details on training of the model, as it falls outside the scope of our work.

While we use the aforementioned model in the generation of the affective dialog, we do not condition the text on any predefined personality.

\subsection{Implementation Details}
\label{ExS}

We conducted experiments with four different affective targets. Marking significant points in the affective space, affective targets with \textit{Valence Arousal Dominance} values of \textit{0.0--0.5--0.0}, \textit{0.0--0.0--0.0}, \textit{0.5--0.0--0.5}, and \textit{1.0--1.0--1.0} were selected for experimentation. The affective target \textit{0.0--0.5--0.0}, which we refer as Low-Medium-Low (LML),  represents sentences that are negative, elicit moderate arousal and are low in dominance, such as sentences including words akin to \textit{stress}, \textit{death}, \textit{torture}. Similarly, the affective target \textit{0.0--0.0--0.0} (LLL),  represents sentences that are negative, low in dominance, but elicit no arousal, such as sentences with words \textit{sad}, \textit{exhausted}, \textit{tiring}. Whereas, \textit{0.5--0.0--0.5} (MLM) affective target is considered the neutral affect, as neutral sentences belong to the middle of the scale in terms of valence and dominance, but induce no arousal (e.g. words close to \textit{0.5--0.0--0.5}: \textit{routine}, \textit{curtain}, \textit{textbook}) \cite{warriner2013norms}. The \textit{1.0--1.0--1.0}, (HHH for High-High-High), affective target is the other extremum, as it signifies sentences that are exceptionally positive, prompt high arousal, and are also dominant, such as those that include words \textit{happiness}, \textit{fun}, \textit{excitement}.

For each experiment, we mapped Cornell Movie Subtitle Corpus \cite{ Danescu-Niculescu-Mizil+Lee:11a} to one of the affective targets. This corpus is a large collection of fictional conversations extracted from movie scripts. It contains a total of 304,713 utterances exchanged among characters.
We experimented with five $\lambda$ values (1.0, 0.7, 0.5, 0.3, 0.0) for each of the affective targets. Note that $\lambda=0$ results in the same output for all of the affective targets, since during decoding no weight is given to affect.
We experimented with $k=20$, $k=30$ and $k=50$, as the number of candidate words.

\section{Evaluation}

We perform subjective and objective evaluation on our proposed approach.

\subsection{Subjective Evaluation}

For the subjective part of the evaluation, the movie corpus was mapped into \emph{HHH}, \emph{MLM}, \emph{LML} affective targets by utilizing the approach proposed in section \ref{AffectON} (with $\lambda=0.5$ and $k=30$, as we observed it to be optimal from objective evaluation). We chose these three targets for subjective evaluation, as the objective results showed them to be distinctive from one another in terms of VADER score. Also we did not collect subjective evaluations for target \emph{LLL}, because from objective evaluations we noticed that sentences mapped into \emph{LLL} changed less than those mapped into \emph{HHH} and \emph{LML}. We were interested to obtain human ratings on how well our approach has mapped sentences that have changed more substantially.

Since not all of the utterances change affect during the mapping, for human evaluation we picked the utterances that were modified the most by calculating n-grams between the mappings and the original utterances. These utterances are most likely to be in the targeted affect, but also to be worse off in terms of perplexity. Other than this criteria, the utterances shown were randomly selected from the three mapped corpora. To avoid any kind of rater bias, we also shuffled the utterances originating from the different corpora when they were presented to the raters.

\subsubsection{Task \& Procedure}

Each rater rated a total of 21 pairs of utterances plus two ``golden`` pairs that were used to filter the ratings high in quality. A \textit{pair} refers to an utterance from the original corpus and its mapped version in one of the affective targets. Utterances were evaluated on a scale from 0 to 100 regarding each affective dimension (i.e. \textit{valence, arousal, dominance}). In addition, they were scored on a scale from 0 to 2 for \textit{syntactic coherence} (Is the utterance grammatically correct?) and \textit{appropriateness} (Is this utterance a plausible reply for the preceding utterance?). In the \textit{appropriateness} part of the evaluation, participants were shown also the preceding dialog utterance, and were asked whether the modified utterance was an appropriate response to the preceding one. Human raters evaluated a total of 273 pairs of utterances, of those 180 distinct pairs of utterances, for the three targets (60 per target).  

\subsubsection{Design and Analysis}
The study was a within-subject design. Each participant was presented both with the original utterance and its mapped version in either of the affective targets. We used one-way ANOVA for analyses.

\subsubsection{Participants}
We recruited participants online via Amazon Mechanical Turk. Participation was limited to adults residing in the United States.
We excluded the data of participants based on their response to two golden questions. These two utterances had opposite valence (e.g. ``she was very happy!'' and ``she was very sad!''). Participants that rated these two utterances as having similar valence were removed from analysis. 13 out of 20 participants who completed the task were retained for final analysis. Each participant was paid 3\$ per task (10\$/hour).
Participants were introduced with concepts of valence, arousal and dominance through a series of explanations and examples prior to proceeding with the task.

\subsubsection{User Interface}
The interface design is regarded crucial to and highly associated with the quality of ratings. Evidently, an inadequate user interface design leads to bias in rating tasks \cite{cosley2003seeing}. We designed a custom web-based interface for the rating task.

Participants rated the utterances, in terms of valence, arousal, dominance, syntactic coherence and appropriateness. Two approaches can be taken in this rating scenario, namely utterance based or dimension based. In the first approach, participants score an utterance in all dimensions (VAD, syntactic coherence, appropriateness) then proceed to the other utterance to score it. The second approach is for participants to score initially all of the utterances in a single dimension, then proceed to the other dimension to score the utterances. A pilot study we conducted indicated that participants had a harder time with the first approach, since changing the mode of the evaluation in addition to increasing the time per rating, also led to confusion among the meaning of dimensions. Hence, we asked participants to score the all utterances in each dimension sequentially.

\begin{figure*}
\includegraphics[width=\linewidth]{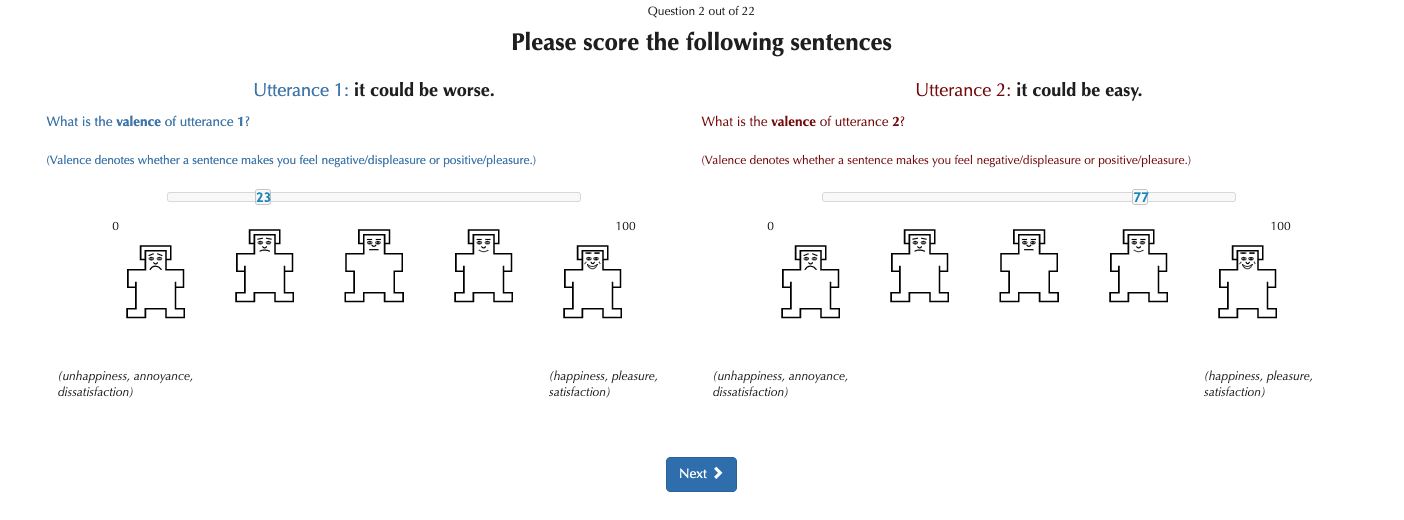}
\caption{The designated user interface for subjective evaluation. Illustrated is a sample question where participants were requested to score the utterances in terms of \emph{valence}. The interface looks similarly for \emph{arousal} and \emph{dominance} dimensions, with respective manikins adopted from \cite{bradley1994measuring}.}
\label{fig:interface}
\end{figure*}

Figure \ref{fig:interface} depicts the user-friendly interface designed for the study. In this example, the rater was asked to score the utterances in terms of valence. Just as shown in the figure, for the dimension that was being scored (i.e. valence) we provided the definition, the manikin adopted from \cite{bradley1994measuring} and the words associated with the extrema of the scale, as an aide-m\'emoire for the raters. Because our affective lexicon has scores on a scale from 0-1, we use a scale from 0-100, as it easier to grasp than decimals, and display it as a slider. Raters were asked to adjust the slider where they thought suitable for the utterance in question. For scoring syntactic coherence and appropriateness, a Likert scale (0-2) like question was presented to raters.

\subsection{Objective Evaluation}

Evaluation of affective language generation systems entails two aspects, namely assessment of how good the generated language is and whether the generated language is in the targeted affect. Below, we discuss the objective metrics used for the two parts of the evaluation. 

\subsubsection{Language Generation}

Generative models in general suffer from lack of established objective evaluation metrics \cite{xie2017neural}. For natural language generation, the most widely reported score is the \textit{perplexity} of the model. Perplexity is a measurement of how well a probability distribution or probability model predicts a sample. A lower \textit{perplexity} indicates that a sentence has higher probability under the language model. Following prior work \cite{Dathathri2020Plug}, we evaluate our generated responses in terms of perplexity with another large language model -- GPT-2 \cite{radford2019language}.

Another broadly established metric in evaluation of conversational models with genesis from machine translation is BLEU (Bilingual Evaluation Understudy) score \cite{papineni2002bleu}. It measures the lexical overlap between the generated responses and the reference ones. Though this metric can be used in cases where there is a reference, or ground truth sentence to be compared to, in our setting we do not have ground truth affective sentences. 
We report BLEU metric with the reference being the original corpus to indicate how different is the affective corpus from the original one. A higher BLEU score indicates fewer utterances were changed in the mapped corpus.

\subsubsection{Sentiment analysis}

While there is no tool available for automatic measurement of affect of a text in terms of VAD, a great deal of effort has been spent on developing automatic tools for sentiment analysis \cite{tausczik2010psychological, hutto2014vader}. Practically, the available tools measure only the valence or polarity of text and do not consider arousal and dominance dimensions. To measure the valence of our generated sentences, we use one of these tools: Valence Aware Dictionary for Sentiment Reasoning (VADER)\cite{hutto2014vader}. VADER is a rule-based system for sentiment analysis. In addition to utilizing a large lexicon of human valence ratings, VADER considers exclamation points, degree modifiers (e.g. \textit{really} good), negation words (e.g. \textit{not} bad), when computing the valence of a given sentence. VADER's compound score classification thresholds are set by the authors \cite{hutto2014vader} at –0.05 and +0.05. If a sentence has a score higher than +0.05 it is considered positive in valence, if it has a score between -0.05 and +0.05 it is considered neutral and if it has a score lower than -0.05 it is considered negative.

\section{Results \& Discussion}

In the following section we discuss the subjective and objective results of our evaluation. Overall, the results suggest that our approach is successful in pulling the generated language towards the targeted affect. Some sample inputs and their responses in different affective targets are provided in Table \ref{table:sample}.

\subsection{Subjective Results}

\begin{figure}[!h]
\subfigure[]{
\includegraphics[width=.5\textwidth]{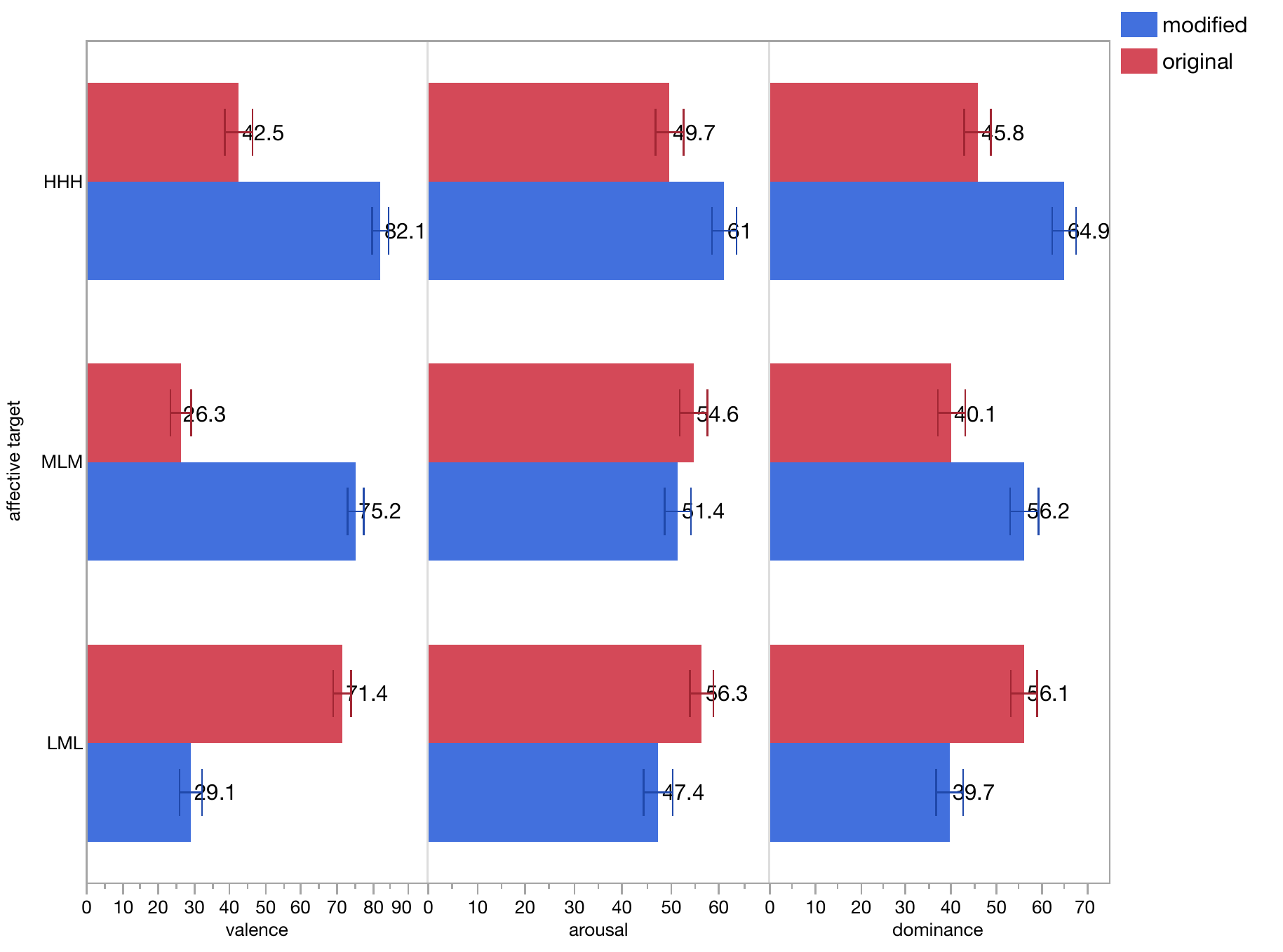}
}

\subfigure[]{
\includegraphics[width=.5\textwidth]{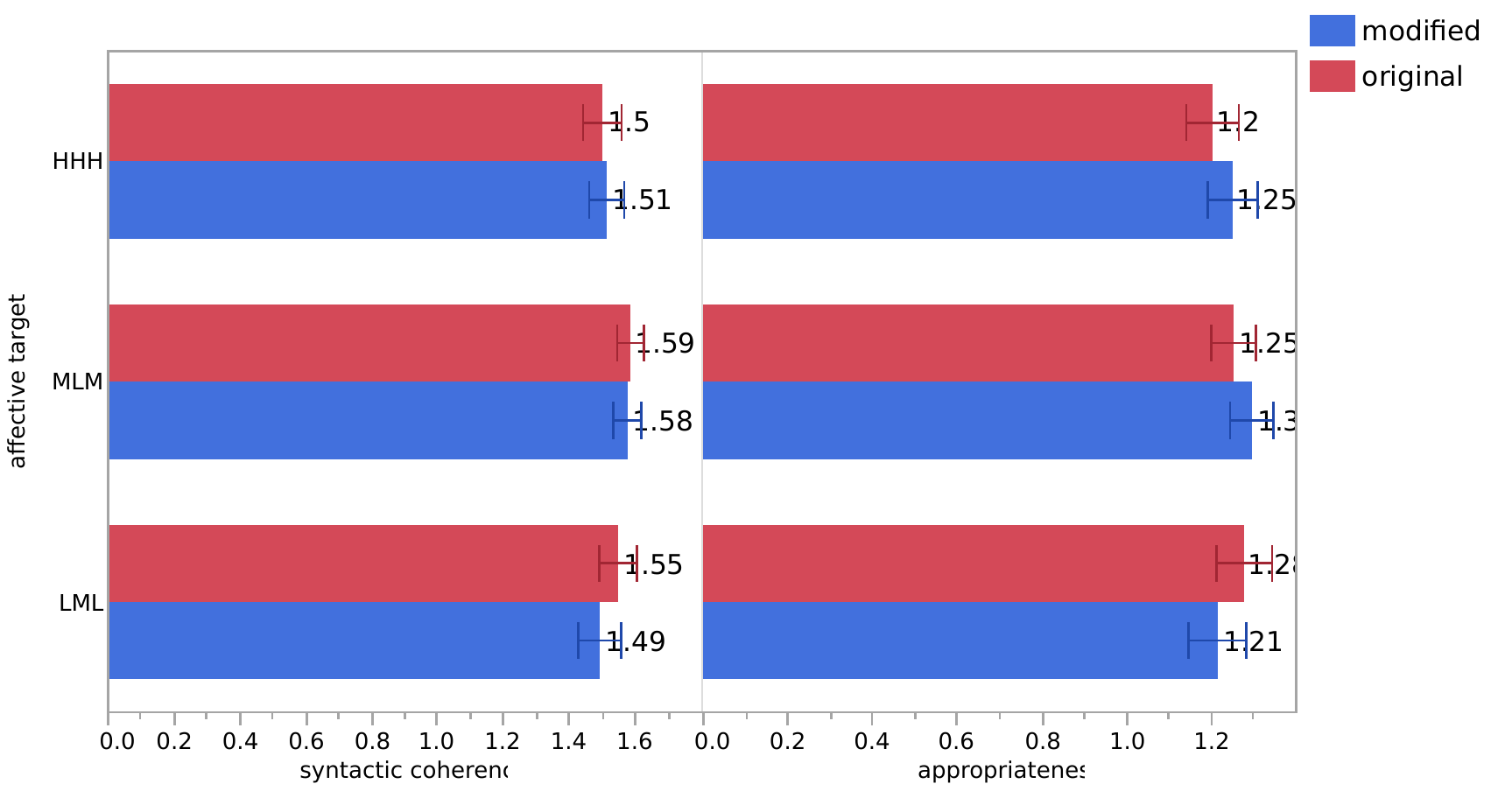}
}

\caption{ Human ratings of (a) valence, arousal and dominance (b) syntactic coherence and appropriatenes for affective targets HHH, MLM and LML (blue) and their original counterparts (red). Error bars indicate one standard error.}
\label{fig:sbjresults}
\end{figure}

The results of the subjective evaluation of utterances in terms of valence, arousal and dominance are shown in Figure \ref{fig:sbjresults} (a). Whereas, Figure \ref{fig:sbjresults} (b) depicts the subjective evaluations of utterances in terms of syntax and appropriateness as a response. For each of the target affects, the mean of the modified sentences and the mean of their original/unmodified counterparts is depicted. Additionally, mean of syntactic coherence and appropriateness are shown. 

We measure the rater agreement for the 186 sentences rated by at least two raters with Pearson correlation for continuous variables (valence, arousal, dominance) and Cohen's $\kappa$ for categorical variables (syntax and appropriateness). Our analysis indicate a very strong correlation for valence (Pearson $r = 0.97, p<<0.0001$), a moderate one for arousal (Pearson $r = 0.42, p<<0.0001$) and a weak one for dominance (Pearson $r = 0.16, p=0.005$). This might be partly due to the fact that arousal and dominance are harder concepts to gauge in comparison to valence. For syntax and appropriateness, raters showed moderate (Cohen's $\kappa$ = 0.52) and fair agreement (Cohen's $\kappa$ = 0.38), respectively.

The results of subjective evaluations indicate that our approach was successful in pulling the mean of valence of the utterances towards the targeted affects. For each of the dimensions, we observed a significant change in \textit{valence}. The valence of original utterances $(M_{o_{HHH}} = 42.47)$ was pulled towards positive affect in \emph{HHH} mapping $(M_{HHH} = 82.09, F_{1,181} = 77.99, p << .0001)$. We observe that valence of the original utterances in \emph{MLM} is negative, whereas valence of \emph{LML} is positive. This is due to the fact that for subjective evaluation, we picked utterances that changed the most, by calculating distinct n-grams between the mappings and the original. These are the utterances that have most likely changed affect in the mapping. For \emph{MLM}, we observe that the modified utterances $(M_{MLM} = 75.17)$ pass $50$, the neutral threshold and have changed significantly from the original utterances $(M_{o_{MLM}} = 26.27, F_{1,181} = 176.10, p << .0001)$. While the result corresponds to the objective VADER score in Figure \ref{fig:results}(b), for $\lambda=0.5$, as discussed more thoroughly in section \ref{quant}, this phenomenon might be a result of sentiment bias of the language model.
Original utterances $(M_{o_{LML}} = 71.38)$ were successfully pulled toward negative valence for affective target \emph{LML} $(M_{LML} = 29.09, F_{1,181} = 112.10, p << .0001)$. 

\begin{figure*}
\centering
\subfigure[Valence]{
\includegraphics[width=.3\linewidth]{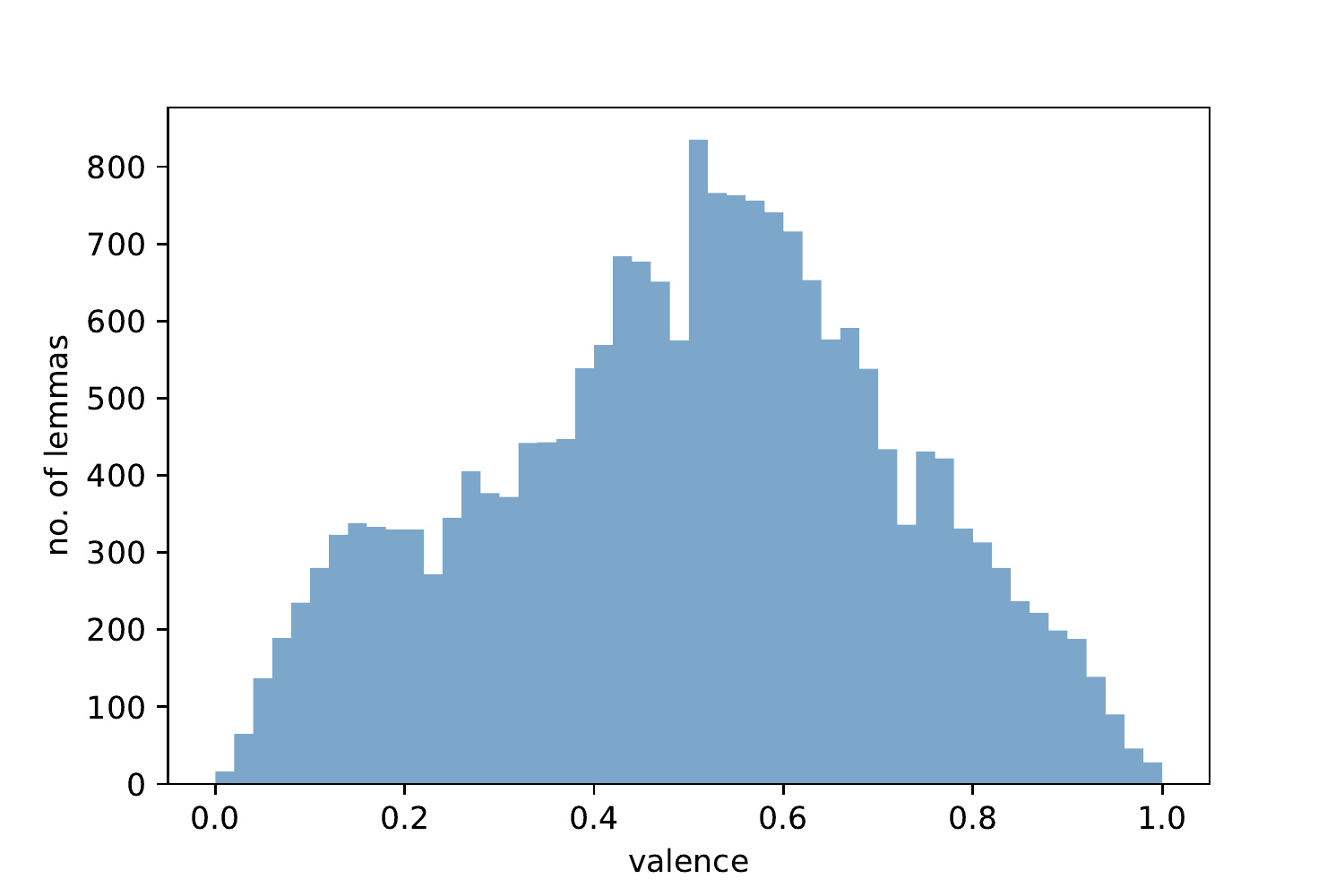}
}
\subfigure[Arousal]{
\includegraphics[width=.3\linewidth]{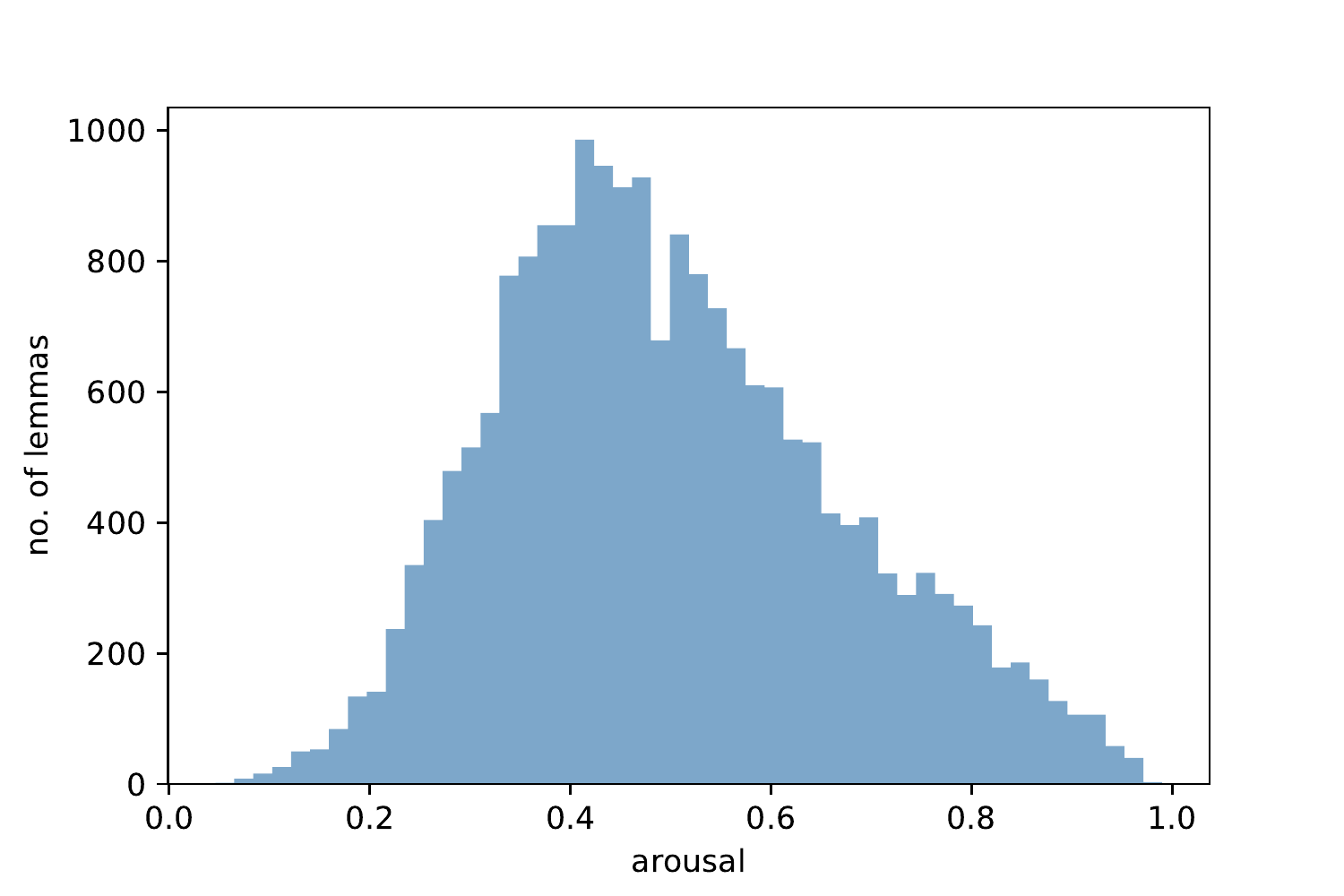}
}
\subfigure[Dominance]{
\includegraphics[width=.3\linewidth]{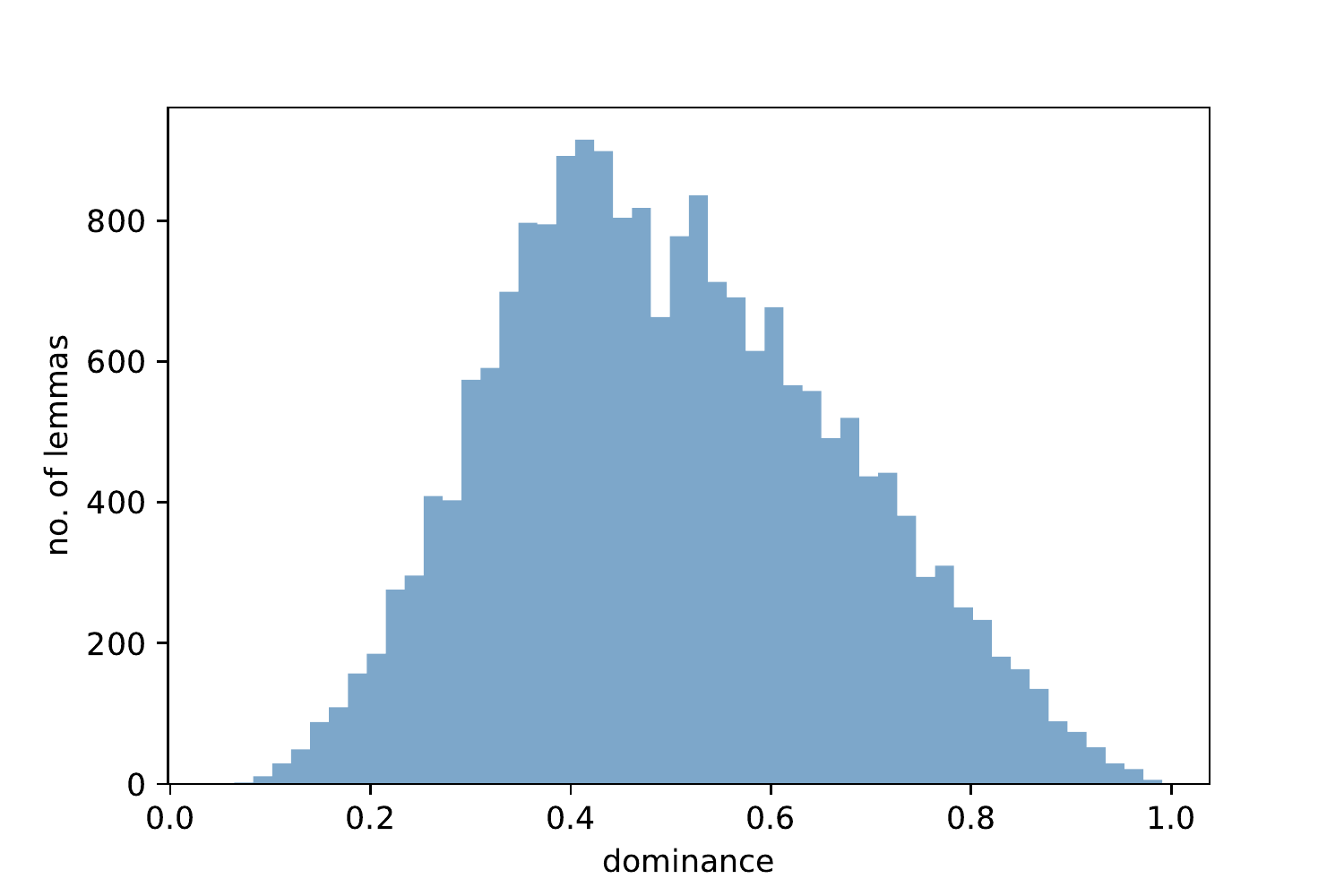}
}
\caption{Distribution of 20,007 lemmas in the affective lexicon according to their VAD values \cite{mohammad2018obtaining}.}
\label{fig:distribution}
\end{figure*}

In contrast to valence, in the distribution of arousal, the words are mostly accumulated in the middle of the scale (See Figure \ref{fig:distribution}). This lack of words high in arousal is perceived in the results of extreme targets $HHH$. While arousal was changed significantly from original $(M_{o_{HHH}} = 49.68)$ towards higher values for target affect \emph{HHH} $(M_{HHH} = 60.95, F_{1,181} = 8.73, p = .003)$, the change was not as perceivable as for valence. Significant change in the negative direction of arousal was also detected for \emph{LML}. Original utterances $(M_{o_{LML}}= 56.28)$ shifted towards low arousal when mapped $(M_{LML} = 47.35, F_{1,181} = 5.37, p =0.02)$. For affective target \emph{MLM}, original utterances $(M_{o_{MLM}}= 54.62)$ moved toward the lower part of the arousal scale $(M_{MLM} = 51.39, F_{1,181} = 0.67, p =n.s.)$, although no significant effect was observed. 

Dominance was pulled significantly in the respective targets for each target affect. For target \emph{HHH}, original utterances $(M_{o_{HHH}} = 45.81)$ are effectively pulled toward high dominance $(M_{HHH} = 64.94, F_{1,181} = 23.76, p << .0001)$. A similar occurrence is also noticed for target dominance target affect $MLM$, where original sentences shift from $(M_{o_{MLM}} = 40.08)$ to $(M_{MLM} = 56.15, F_{1,181} = 13.62, p = .0003)$. For target \emph{LML}, utterances shifted from $(M_{o_{LML}} = 56.07)$ toward lower dominance $(M_{LML} = 39.65, F_{1,181} = 15.58, p << .0001)$, as desired. 
 
We did not observe any significant effect in terms of difference in syntactic coherence between the modified sentences and the original ones, for either any of the target affects: \emph{HHH} $(M_{o_{HHH}} = 1.50, M_{HHH} = 1.51, F_{1,181} = 0.61, p=n.s.)$, \emph{MLM} $(M_{o_{MLM}} = 1.59, M_{MLM} = 1.58, F_{1,181} = 0.56, p=n.s.)$, \emph{LML} $(M_{o_{LML}} = 1.55, M_{LML} = 1.49, F_{1,181} = 0.62, p=n.s.)$. However, the mean of both the modified and original utterances is close to 1.5 on a scale from 0 (not grammatically correct) to 2 (grammatically correct) which indicates that the original utterances, also, were not syntactically coherent to begin with. We observe that our model preserves the syntactic coherence. For example, original \emph{MLM} utterances, have higher syntactic coherence and their modified counterparts also stay in the same range. As seen in the objective evaluation, affective target \emph{LML} disturbs perplexity the most among other targets. This is also reflected in the subjective evaluations.

Since the dialog pairs originate from a movie subtitle corpus, they generally lack the naturalness of daily conversation. This unnaturalness is reflected on the \textit{appropriateness} score, where even the original dialog pairs have a relatively low mean of 1.2 on a scale from 0 (inappropriate response) to 2 (appropriate response). As it is the case with syntactic coherence, the utterances are more appropriate when the original affect and the targeted affect are not very different (i.e. affective target is $MLM$). However, we do not observe significant difference in terms of appropriateness of original and modified utterances for any of the affective targets: \emph{HHH} $(M_{o_{HHH}} = 1.20, M_{HHH} = 1.25, F_{1,181} = 0.24, p=n.s.)$, \emph{MLM} $(M_{o_{MLM}} = 1.25, M_{MLM} = 1.3, F_{1,181} = 0.33, p=n.s.)$, \emph{LML} $(M_{o_{LML}} = 1.28, M_{LML} = 1.21, F_{1,181} = 0.24, p=n.s.)$

Overall, our results indicate that our approach is mostly successful in pulling the generated language towards the target affect, without any significant loss in terms of syntactic coherence and appropriateness. 

\subsection{Objective Results}
\label{quant}

\begin{figure}
\subfigure[$k=20$]{
\includegraphics[width=.5\textwidth]{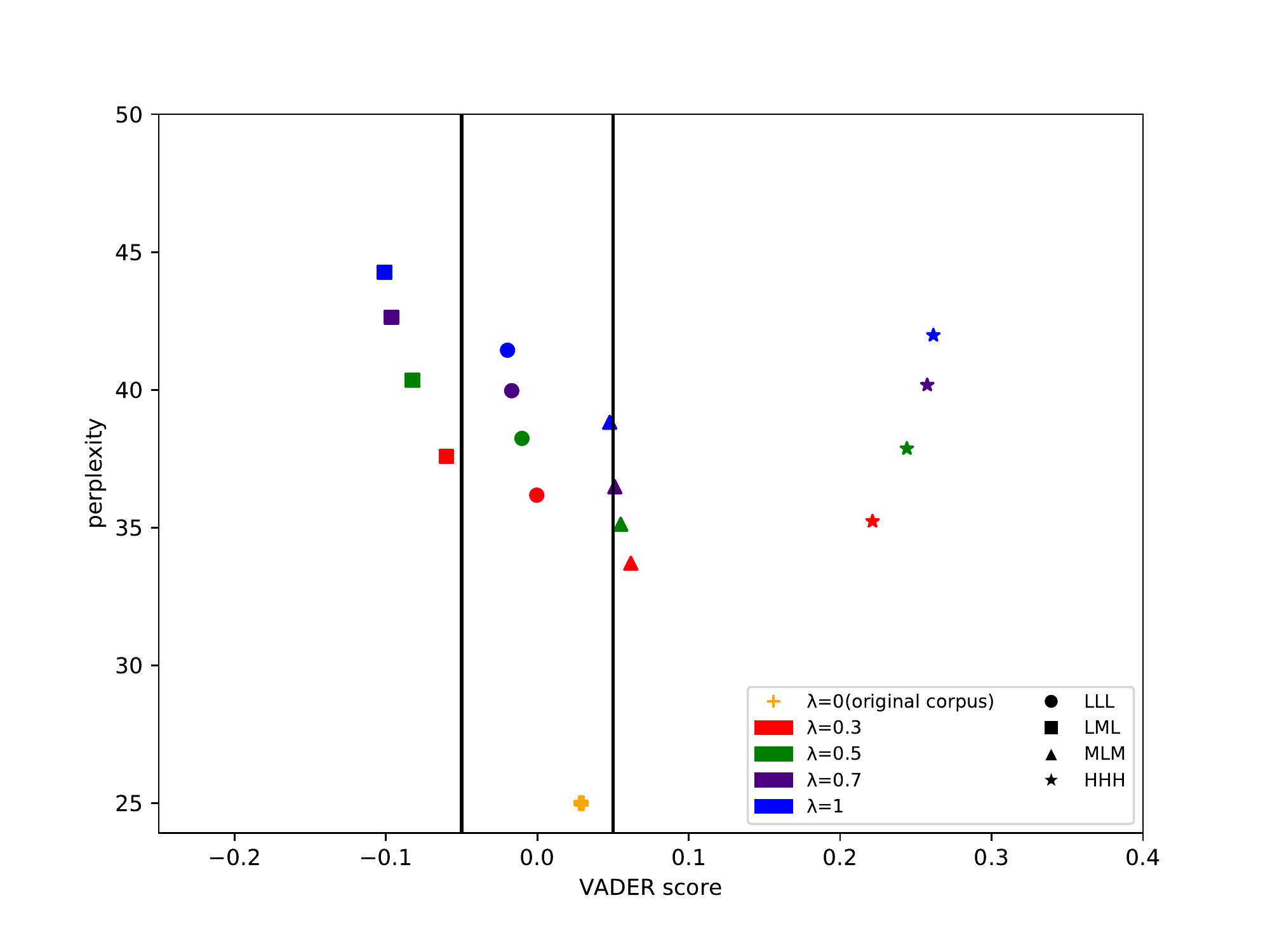}
}
\subfigure[$k=30$]{
\includegraphics[width=.5\textwidth]{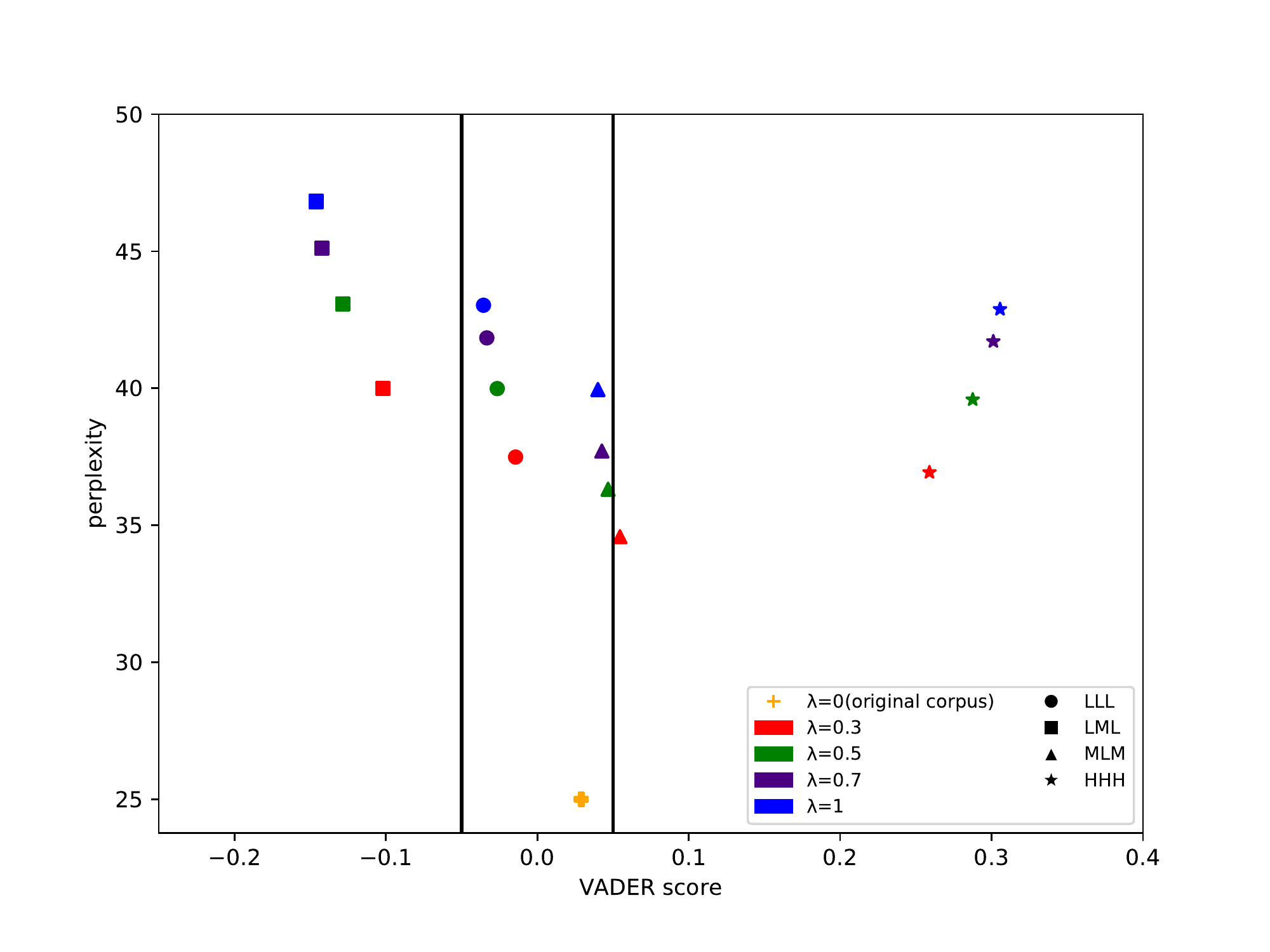}
}
\subfigure[$k=50$]{
\includegraphics[width=.5\textwidth]{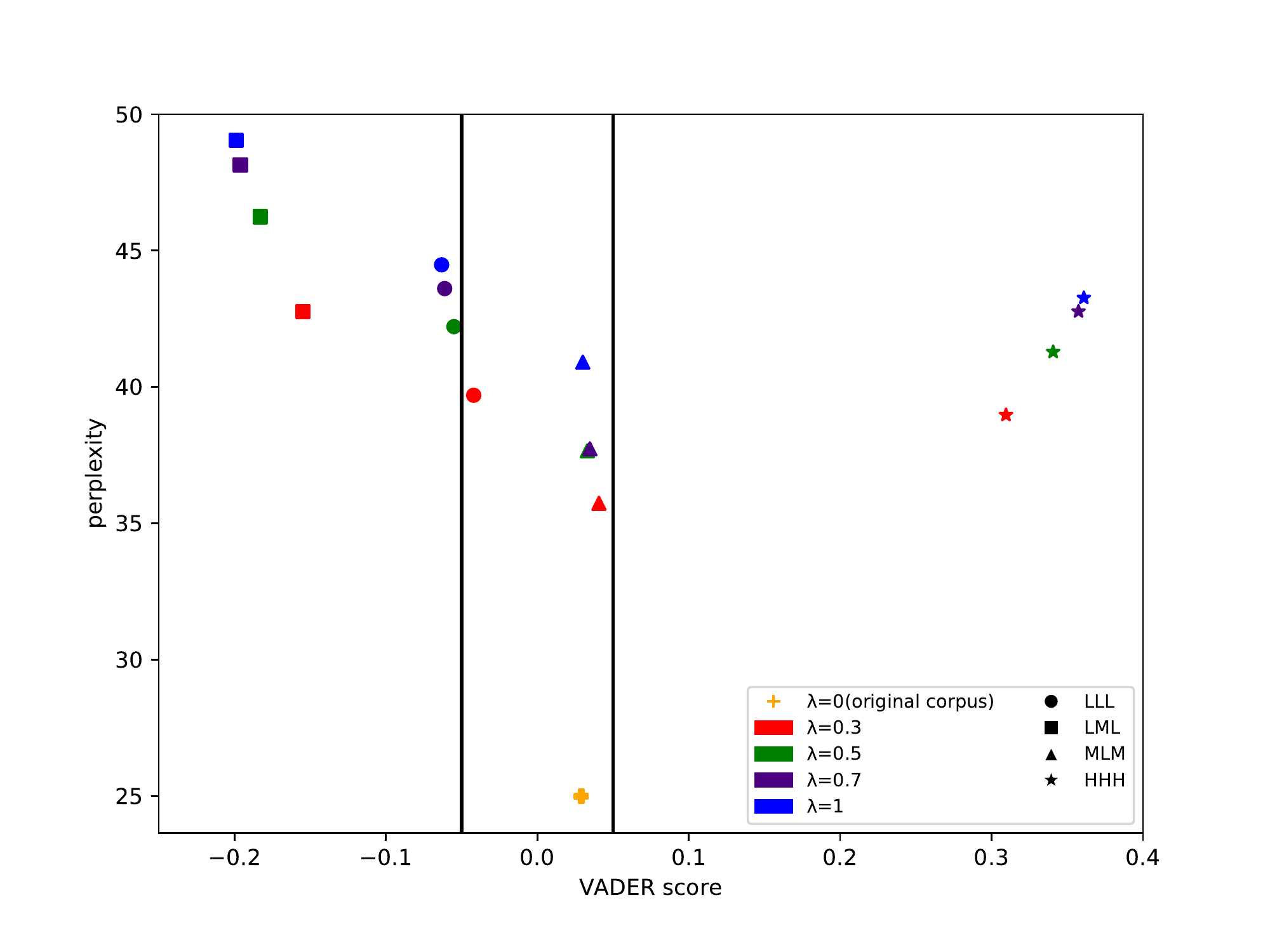}}

\caption{Results of experiments with different number of candidate words $k$ and $\lambda$ values. Each point represents the average VADER score and the perplexity of the movie corpus mapped to a given target affect. The two vertical lines depict the VADER score boundaries for positiveness ($>= 0.05$) and negativeness ($=< -0.05$).}
\label{fig:results}
\end{figure}

Figure \ref{fig:results} depicts the averages over the entire movie corpus (304, 713 utterances) for $k=20$, $k=30$ and $k=50$. Comparing the results of different $\lambda$ values for each $k$, we observe that the larger the $\lambda$ value, the less importance is given to the language model constraint, hence the wider range of VADER scores (valences). However, as expected the perplexity worsens as $\lambda$ increases. We observe that when $\lambda$ is small (e.g. $\lambda = 0.3$), perplexity scores are better than for larger $\lambda$ values, but the range of VADER scores is very narrow. Thus, some of the affective targets do not exceed their respective VADER score boundary due to the priority that is given to language model. Perplexity scores imply that our approach performs the affect modification without a huge deterioration in language structure for smaller $\lambda$. Among modifications with the same $\lambda$ values but different target affects, again we observe that targets closer to original affect have better scores.

Comparing the results of different $k$ values, we notice that as we increase number of candidate words we see better mappings for valence. For $k=50$, except when $\lambda=0.3$, \emph{LLL} falls in the negative valence portion, and the neutral affect \emph{MLM} falls in the neutral zone, as desired. However, the trade-off is seen in terms of perplexity. Evidently, increasing the number of candidate words allows the algorithm to select a word from a larger pool of words, which might be closer to the target affect, yet has lower probability under the language model.

Among these experiments, AffectON with $k=30$ and $\lambda=0.5$ manages to modify the affect noticeably, yet preserve the perplexity scores the most. We expected the original corpus ($\lambda=0$) utterances and the modified utterances of $\lambda=0.5$ with the neutral target affect $a_t = [M, L, M]$ to have the similar average VADER scores. However, in Figure \ref{fig:results}(a) we see that $\lambda$ (0.3, 0.5) values of the neutral affective target fall in the positive side of the VADER border. We observe a similar effect for the \emph{LLL} affective target, for which the values fall in the neutral zone of the sentiment, instead of the negative. This is particularly interesting because it suggests that the underlying language model is biased towards positive sentiment, in line with evidence from prior work \cite{arnold18:sentiment}.

We also observe that mapping sentences to the neutral affet \emph{MLM}, as expected, has the least negative impact on perplexity. We notice that target \emph{LML} has the highest perplexity, although the distance between the original corpus and the mappings in terms of valence is smaller than that of mappings for affective target \emph{HHH}. This again highlights the bias of language model towards the positive valence as it gives positive words higher probabilities.

Finally, to understand how distinct the mappings are from the original corpus we computed BLEU scores (computed using 1- to 4-grams) with the original corpus as the reference. BLEU scores, presented in Table \ref{bleu}, communicate a similar story as perplexity. As the number of candidate words increases, BLEU score decreases, indicating that the mappings become more different from the original corpus. We see a similar pattern with $\lambda$ values, also. Higher $\lambda$ values allow the model to pick words that are more affective, and in turn less similar to the original corpus, resulting in lower BLEU scores.

\subsection{Correlation between Objective Metrics and Human Judgments}

To understand how reliable VADER and perplexity are as objective measures for our task, we computed the correlations between these two metrics and human judgements of valence and syntactic coherence, respectively. In line with prior work that analyzes the correlation between objective metrics and human judgments for natural language generation \cite{reiter2009investigation}, we compute these correlations on individual ratings of utterances and also their mean values. Correlations amongst mean values are computed because in natural language generation, metrics (e.g. perplexity, BLEU score) are more informative when reported on corpus level (i.e. averages) rather than per sentence \cite{reiter2009investigation}. However, the number of data points is reduced notably when computing correlations on mean values, thus significance cannot be shown. 

We picked $k=30$ and $\lambda=0.5$ for the subjective evaluation, hence we compare mean values of VADER score and perplexity of that particular setting to mean human ratings. These scores are shown in Table \ref{objsbj}.
The results revealed a strong and highly significant correlation between 546 individual ratings of valence and VADER scores of the same utterances (Pearson $r = 0.73, p<<0.0001$). Note that these 546 utterances are the 273 rated utterances and their corresponding ground truth pairs. We observed a strong correlation between means of VADER and valence scores of affective targets, also (Pearson $r = 0.88$).

When measuring the correlation between perplexity and syntactic coherence on individual 506 ratings, we used one-way ANOVA, because syntactic coherence is a categorical variable for individual ratings. The results of the analysis were not significant ($F_{1, 2}=1.44, p=n.s.$). However, we noticed a strong correlation between the mean values of perplexities and syntactic coherence ratings per affective target (Pearson $r = -0.92$). Note that the correlation is negative because higher perplexity indicates a less probable sequence.

Overall, VADER seems to be a good predictor of human ratings of valence. We investigated the reasons behind not finding a significant effect between perplexity and syntactic coherence for individual ratings. Particularly, we looked into sequences where there was discrepancy between the two ratings. These included sequences that human rated as syntactically incoherent, while they had low perplexities under the language model. Examples included sequences such as: ``He's a pretty man.'', where other adjectives are more commonly used to describe the subject (e.g. \emph{handsome}). Our observations suggest that perplexity is a relatively good gauge of the overall plausibility of sequences, especially when calculated on corpus level. However, subjective evaluation is still necessary to measure the quality of the generated language.

\begin{table}[]

\centering
\caption{Objective Measures and Human Judgments ($k=30$, $\lambda=0.5$)}
\label{objsbj}
\begin{tabular}{l|ll|ll}
\textbf{}    & \textbf{VADER} & \textbf{valence} & \textbf{perplexity} & \textbf{syntactic coherence} \\ \hline
\textbf{HHH} & 0.28           & 82.1               & 39                  & 1.51                         \\
\textbf{MLM} & 0.04           & 75.2               & 36                  & 1.58                         \\
\textbf{LML} & -0.13          & 29.1               & 43                  & 1.49                        
\end{tabular}
\end{table}

\begin{table}[]
\centering
\caption{BLEU scores for $k=20$, $k=30$ and $k=50$ candidate words. Each number indicates the BLEU score of the mapped corpus.\label{bleu}}

\begin{tabular}{@{}lllll@{}}
\toprule
\textbf{$k=20$}   & $\lambda$ =1.0 & $\lambda$ =0.7 & $\lambda$ =0.5 & $\lambda$ =0.3 \\ \midrule
HHH & 0.079          & 0.081          &  0.085          & 0.090          \\
MLM & 0.086          & 0.092          & 0.096          & 0.097           \\
LML & 0.086          & 0.089          & 0.093         & 0.096           \\
LLL & 0.086          & 0.089          & 0.094          & 0.097           \\ \bottomrule

\end{tabular}

\begin{tabular}{@{}lllll@{}}
\toprule
\textbf{$k=30$}   & $\lambda$ =1.0 & $\lambda$ =0.7 & $\lambda$ =0.5 & $\lambda$ =0.3 \\ \midrule
HHH & 0.071          & 0.073          & 0.077          & 0.082          \\
MLM & 0.077          & 0.084          & 0.088          & 0.090           \\
LML & 0.074          & 0.078          & 0.082          & 0.087           \\
LLL & 0.076          & 0.079          & 0.084          & 0.088           \\ \bottomrule

\end{tabular}

\begin{tabular}{@{}lllll@{}}
\toprule
\textbf{$k=50$}   & $\lambda$ =1.0 & $\lambda$ =0.7 & $\lambda$ =0.5 & $\lambda$ =0.3 \\ \midrule
HHH & 0.063          & 0.067          & 0.071          & 0.077          \\
MLM & 0.065          & 0.071          & 0.081          & 0.085           \\
LML & 0.062          & 0.070          & 0.074          & 0.078           \\
LLL & 0.069          & 0.071          & 0.075          & 0.081          \\ \bottomrule

\end{tabular}

\end{table}

\begin{table*}[]
\centering
\caption{Sample inputs and their responses in different affective targets. \label{table:sample}}
\begin{tabular}{@{}ll@{}}
\toprule
\textbf{input} & Good to see you!                                        \\ \midrule
\textbf{target VAD - MLM}   & It's been a long day. How are you?                                     \\
\textbf{target VAD - HHH}   & It's been a wonderful pleasure chatting with you! How are you doing right now?                                   \\
\textbf{target VAD - LML}   & It’s been a crazy day! How are you doing?                                \\
\textbf{target VAD - LLL}   & It's been a little lonely here in the wilderness.                                       \\ \midrule
\textbf{input} & Maybe we could go somewhere and just eat?       \\ \midrule
\textbf{target VAD - MLM}   & I'm not sure. \\
\textbf{target VAD - HHH}   & I'm great at eating. I'm very good at it. \\
\textbf{target VAD - LML}   & I'm not hungry. You can go.     \\
\textbf{target VAD - LLL}   & I'm not one for small talk. I'm a loner.    \\ \midrule
\textbf{input} & Can you help us carry this?                     \\ \midrule
\textbf{target VAD - MLM}   & I can be of use to you.                     \\
\textbf{target VAD - HHH}   & I love to help people. I'm a fierce leader.                      \\
\textbf{target VAD - LML}   & I need to leave. I don’t have time.                     \\
\textbf{target VAD - LLL}   & I suppose so. I'm not used to people helping me, though.             \\ \bottomrule
\end{tabular}
\end{table*}

\section{Conclusion}

As conversational agents have become ubiquitous, it is paramount that our interaction with them is effortless and satisfying. This also means that these conversational agents need to be affectively cognizant and respond to our affective states appropriately.
In this paper, we presented AffectON, an approach for generating affective language. Particularly, we experimented with affective dialog generation. We leveraged a sequence-to-sequence language model and an affective space to generate affective responses. We conducted subjective and objective evaluations to assess our approach. Our results indicate that AffectON can successfully pull text to a targeted affect, with a small sacrifice in terms of syntax. Our results also indicate that the objective metrics correlate reasonably well with subjective ratings.
With the rapid pace of development of large language models, AffectON is promising as an approach that generates affective responses relatively well while intervening only during inference. Therefore, it can easily accommodate any underlying language model without the need for further training.

Despite its promises for affective language generation, this study has some limitations. For instance, negation words (e.g. \textit{not}) are an issue, since we apply our approach word by word. In a positive target affect, for a instance, the original sentence \textit{The movie wasn't bad.} might become \textit{The movie wasn't good}. Though, the objective evaluation (VADER score) considers these negation words and the results show that our approach is successful in pulling the language toward the desired affect, in this context there is room for improvement.

While we present an approach for generating affective language, further research is necessary to understand affective dialog strategies. For example, when should the conversational agent match user's affective state and when it is beneficial to generate responses in a different affect? We should also look deeper into human-human interactions and understand which strategies work best for which types of situations, applications and people.

\section{Acknowledgements}
Authors M. Sezgin, Y. Yemez and E. Erzin thank ERA-Net CHIST-ERA (JOKER project) and The Scientific and Technological Research Council of Turkey (TUBITAK, grant number 113E324).


%

\ifCLASSOPTIONcaptionsoff
  \newpage
\fi



\bibliographystyle{IEEEtran}
%
\bibliography{reference}



%

\begin{IEEEbiography}[{\includegraphics[width=1in,height=1.25in,clip,keepaspectratio]{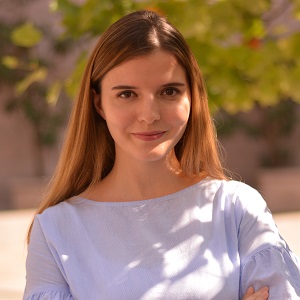}}]{Zana Bu\c cinca} is a PhD student at Harvard University. She received her M.S. degree in Computer Science and Engineering from Koc Univeresity in 2019. She graduated \emph{summa cum laude} with a B.S. degree in Computer Engineering from Izmir Institute of Technology in 2016. Her research interests include human-computer interaction, machine learning and affective computing.
\end{IEEEbiography}
\begin{IEEEbiography}[{\includegraphics[width=1in,height=1.25in,clip,keepaspectratio]{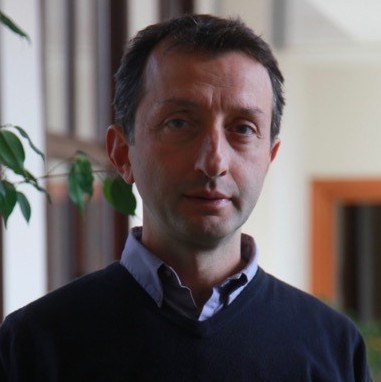}}]{Y\" ucel Yemez} received the BS degree from Middle
East Technical University, Ankara, in 1989 and the
MS and PhD degrees from Bogazici University,
Istanbul, respectively, in 1992 and 1997, all in
electrical engineering. From 1997 to 2000, he was
a postdoctoral researcher in the Image and Signal
Processing Department of Telecom Paris (ENST).
Currently, he is an associate professor in the Computer Engineering Department, Koc University,
Istanbul. His research interests include various
fields of computer vision and graphics.
\end{IEEEbiography}
\begin{IEEEbiography}[{\includegraphics[width=1in,height=1.25in,clip,keepaspectratio]{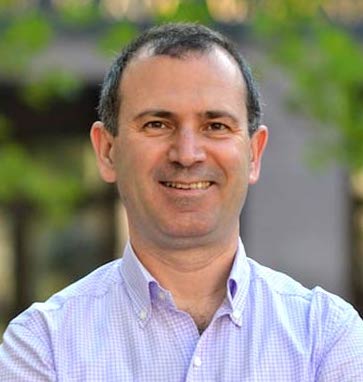}}]{Engin Erzin}
(S’88-M’96-SM’06) received his Ph.D. degree, M.Sc. degree, and B.Sc. degree from the Bilkent University, Ankara, Turkey, in 1995, 1992 and 1990, respectively, all in Electrical Engineering. During 1995-1996, he was a postdoctoral fellow in Signal Compression Laboratory, University of California, Santa Barbara. He joined Lucent Technologies in September 1996, and he was with the Consumer Products for one year as a Member of Technical Staff of the Global Wireless Products Group. From 1997 to 2001, he was with the Speech and Audio Technology Group of the Network Wireless Systems. Since January 2001, he has been with the  Electrical \& Electronics Engineering and Computer Engineering Departments of Koc University, Istanbul, Turkey.  Engin Erzin is currently a member of the IEEE Speech and Language Processing Technical Committee and Associate Editor for the IEEE Transactions on Multimedia, having previously served as Associate Editor of the IEEE Transactions on Audio, Speech \& Language Processing (2010-2014). His research interests include speech-audio-visual signal processing, affective computing, human-computer interaction and machine learning.
\end{IEEEbiography}

\begin{IEEEbiography}[{\includegraphics[width=1in,height=1.25in,clip,keepaspectratio]{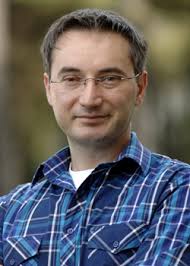}}]{Metin Sezgin} received the graduate (summa
cum laude) degree with Honors from Syracuse University, in 1999. He received the MS degree from the Artificial Intelligence Laboratory, Massachusetts Institute of Technology, in 2001. He received the PhD degree in 2006 from Massachusetts Institute of Technology. He subsequently moved to University of Cambridge, and joined the Rainbow Group at the University of Cambridge Computer Laboratory as a postdoctoral research associate. He is currently an associate professor in the College of Engineering, Koc University, Istanbul. His research interests include intelligent human-computer interfaces, multimodal sensor fusion, and HCI applications of machine learning. He is particularly interested in applications of these technologies in building intelligent pen-based interfaces. His research has been supported by international and national grants including grants from DARPA (USA), and Turk Telekom. He is a recipient of the Career Award of the Scientific and Technological Research Council of Turkey.
\end{IEEEbiography}




\end{document}